# A Data Mining-Based Dynamical Anomaly Detection Method for Integrating with an Advance Metering System

Sarit Maitra
Alliance Business School, Alliance University, Bengaluru, India
sarit.maitra@gmail.com

*Abstract*—**Building operations consume 30% of total power consumption and contribute 26% of global power-related emissions. Therefore, monitoring, and early detection of anomalies at the meter level are essential for residential and commercial buildings. This work investigates both supervised and unsupervised approaches and introduces a dynamic anomaly detection system. The system introduces a supervised Light Gradient Boosting machine and an unsupervised autoencoder with a dynamic threshold. This system is designed to provide real-time detection of anomalies at the meter level. The proposed dynamical system comes with a dynamic threshold based on the Mahalanobis distance and moving averages. This approach allows the system to adapt to changes in the data distribution over time. The effectiveness of the proposed system is evaluated using real-life power consumption data collected from smart metering systems. This empirical testing ensures that the system's performance is validated under real-world conditions. By detecting unusual data movements and providing early warnings, the proposed system contributes significantly to visual analytics and decision science. Early detection of anomalies enables timely troubleshooting, preventing financial losses and potential disasters such as fire incidents.**

I. **INTRODUCTION**

The OECD[1] predicts that the world economy will require 80% more power in 2050 without a new measurement policy. Global power consumption is increasing at an alarming rate, and buildings (residential and commercial) account for approximately 40–45% of global power consumption ([1]; [2]; [3]; [4]; [5]). According to the IEA[2], while global growth in electricity demand will decrease slightly to 2.2% by 2023, it is projected to accelerate to an average of 3.4% from 2024 through 2026. "*The operations of buildings account for 30% of global power consumption and 26% of global power-related emissions (8% being direct emissions in buildings and 18% indirect emissions from the production of electricity and heat used in buildings)."* [3]. Global warming and heatwaves are leading to increased power consumption in terms of air conditioning. In addition, because of the pervasive misuse of residential power consumption behaviors, it is estimated that 15–30% of the power utilized during building operations is lost owing to malfunctioning equipment, poor operation protocols, and poor construction design ([6]; [3]). Reducing electricity usage in buildings could greatly aid in meeting the urgent need to reduce global power consumption and its associated environmental benefits. Although smart metering systems capture data using a combination of technical, environmental, human, and operational factors ([7]; [8]; [9]; [10]), these systems must be integrated with a real-time anomaly detection and alert mechanism. Researchers agree that comprehending a building's power usage behavior both now and, in the future, can help avoid problems with power supply or wasteful use in the medium and long run (Serrano-Guerrero, 2020).

Anomalies refer to data points that significantly deviate from the rest of the dataset. Although these are minority datasets, they have the potential to have serious implications if not immediately identified and root-cause analysis is performed. Efficient anomaly detection is one of the best ways to mitigate waste during building operations [3]; Lei et al., 2023. A clear mathematical definition cannot be applied to explain anomalies since their detection is a more subjective decision than a routine assessment. Therefore, a real-life detection method always prefers to look for a real-time detection system that matches the semantics of the specific application context.

---

[1] https://shorturl.at/dBEX0
[2] https://shorturl.at/rGKN6
[3] https://www.iea.org/energy-system/buildings

Smart electric meters are replacing traditional electric meters as technology development progresses. These are electronic devices that provide data to electricity suppliers and consumers, including voltage levels, current, power factors, and electrical use. In contrast to automatic meter reading, this kind of advanced metering infrastructure (AMI) allows for two-way communication between the supplier and the meter. The deployment of smart meters in smart cities has led to a paradigm shift in power management [7]. These digital devices, which measure and record consumption in real time, are rapidly becoming popular and have opened avenues for data mining. Research on anomaly detection for power consumption has gained momentum in recent years (e.g., [14], [15], [16]). However, the nondeterministic nature of sensor data, coupled with the influence of external factors, often introduces anomalies in time-series data (Jia et al., 2019). These anomalies can significantly impact the reliability and accuracy of automated decision-making, particularly when statistical methods are used. The AMI monitors, gathers, and analyzes power consumption while also communicating with smart devices in real time. It comprises hardware, software, communications, consumption display, and meter data management software.

Our goal is to lower a building's power-related costs by determining how to automatically identify anomalies in near real time, such as unusual power consumption statistics that significantly deviate from usual behavior. To meet this goal, this work provides an empirical investigation of the data collected from smart metering systems and develops a real-time anomaly detection system. Development and implementation involve a series of decisions guided by the principles of decision-making and decision science. These include problem identification and definition, data collection and analysis, risk assessment and management, technology evaluation and selection, algorithm selection and optimization, resource allocation, and continuous improvement and adaptation. Problem identification, which involves recognizing the significant amount of power lost due to anomalies in building operations, is key. Data-driven insights are crucial in identifying patterns of power consumption, which would identify anomalies and inform the detection system to generate alerts.

Researchers typically classify anomalies into three categories: point, contextual, and collective anomalies (Foorthuis, 2021). Our focus here is on contextual and point anomalies. A point in the data separated from the rest is known as a point anomaly or outlier. A point that is normal in one scenario but odd in another is known as a contextual anomaly or conditional anomaly (Luna-Romero et al., 2024). In power monitoring, contextual anomaly detection is a relatively new and unexplored challenge. An early alert would help reduce anomalies and prevent financial losses. Various studies have reported the potential of employing artificial intelligence (AI) to detect anomalous patterns ([17]; [15]; Nassif et al., 2021). Despite the potential of AI, real-time power consumption data do not contain annotations or labels that specify when anomalies occur [18]. Thus, unsupervised ML approaches are favored in this dataset. However, this approach has its own set of challenges, which may include appropriate feature representations, selecting suitable algorithms, addressing data noise and variability, and interpreting detected anomalies. Nassif et al. (2022) conducted an extensive literature review and advised carrying out additional ML studies of anomaly detection to gather more data regarding the effectiveness and performance of ML models.

Table I presents the identified research gap vis-à-vis the key contribution of our work.

TABLE I. RESEARCH GAPS AND KEY CONTRIBUTIONS.

| Research gaps | Our contributions |
|---|---|
| Despite extensive research and significant advances, the discipline of anomaly detection is yet to reach maturity. It lacks an overall, integrative framework to comprehend its nature and its core idea, the abnormality (Foorthuis, 2018). Although the literature on data mining, artificial intelligence, and statistics provides many methods for distinguishing between distinct types of anomalies, research has yet to produce comprehensive and clear overviews and conceptualizations (Foorthuis, 2021). Even though numerous anomaly detectors have been offered over the years, deploying them to a particular service continues to be quite difficult and necessitates manually and repeatedly adjusting thresholds and detector parameters (Liu et al., 2015). | We present a multivariate analysis where an anomaly is assigned to the relationship between its variables. Our work considers the joint distribution to consider the relationships between different variables and their combinations of values. Therefore, our work considers the cardinality of the relationships among selected variables to provide a comprehensive analysis of anomaly detection. Our work presents a dynamical system with the flexibility to apply in both supervised and unsupervised learning environments, combined with a dynamic threshold that is suited for different types of anomalies and data characteristics. The framework provides a unified approach, leveraging the strengths of different methods. |
| Research on building power consumption is not adequate; however, it is showing rapid progress. The number of publications has grown significantly (Han and Wei, 2021). The field is dominated by academic scholars and lacks the perspective of industry practitioners. Although the existing researchers covered diverse fields, data-driven anomaly detection using multivariate analysis in this domain is not prevalent in real life. Most of the existing work used normal distributions for anomaly classification (see Liu and Nielson, 2018). Although the statistical method is effective, it must be used with other advanced techniques for better efficiency. Moreover, researchers investigated ML for anomaly detection in power consumption (see [11]; [12]); however, their studies pertain to fraud detection and do not provide any real-time reporting. Moreover, though scalable anomaly detection for smart meters has been investigated (see [13]), the work has not been evaluated to detect a wide range of anomalies, including missing values, negative power consumption, and device errors. | This study presents a combination of advanced data mining and machine-learning (ML)-based techniques. It combines statistical and both supervised and unsupervised approaches to provide a near-real-time anomaly detection system. The work proposes a dynamic thresholding approach based on Mahalanobis distance (MD) and moving averages (MAV) and is tested with real-life data obtained from a smart metering system. By focusing on real-world deployment and addressing practical challenges faced in smart meter data analysis, such as missing values and device errors, our approach aims to bridge the gap between research and practical implementation. This makes the approach more relevant and applicable in real-world scenarios, where accurate and reliable anomaly detection is crucial for efficient power management |
| In other cases, the whole pattern of power consumption—rather than simply a few isolated observations—may be considered anomaly. This prevents the true anomalous points from being identified. Therefore, rather than depending exclusively on anomaly identification algorithms which find abnormalities at the sampling level, current consumption footprints need to be compared to ideal and historical consumption cycles. The current research on this topic does not clearly offer an empirical analysis. | This study aims to bridge the gap between research and practical implementation by considering past consumption footprints and historical trends to provide a thorough understanding of anomalies. This work considers past consumption footprints and historical trends to provide a thorough understanding of anomalies. By comparing current consumption patterns with historical data, including consumption cycles, this study identifies the deviations. This provides a thorough understanding of anomalies, considering context and historical trends. |

With the advancement of technology, many systems are currently available for collecting and storing power usage data, such as Building Energy Consumption Monitoring Platforms (BECMP), Building Energy Management Systems (BEMSs), and the Internet of Building Energy Systems (IBESs). However, anomalous power consumption data are not always precise, primarily due to Building activities use a lot of energy; they account for over one-third of global energy use (Fan et al., 2018). weather conditions, malfunctioning devices, and the inadequacy of consumption monitoring technology platforms. Moreover, the lack of power models in machine learning frameworks, such as TensorFlow, Caffe2 (Jia et al., 2014), and PyTorch (Paszke, 2017), to support energy evaluations and the lack of familiarity with current methods for estimating energy are the reasons why the machine learning community has not shown increased interest in energy consumption (Garcia-Martin et al., 2019).

This work presents the current state-of-the-art approaches for estimating power consumption by combining data mining and advanced machine learning techniques. The proposed models comprise multivariate analysis with large amounts of data and are presented in both supervised and unsupervised formats to facilitate building power consumption and anomaly detection. Anomaly detection techniques can be categorized as either supervised learning or unsupervised learning (Fan et al., 2018), depending on the data extraction, transformation, and loading (ETL) pipeline. This work investigates both approaches and proposes a dynamical system based on a dynamic threshold.

## II. DATA DESCRIPTION ANALYSIS

The data include hourly power measurements of three factory monitoring systems published by Schneider Electric[4] and span the years 2012–2016. It is supported by three additional datasets: (1) metadata offering location information and general descriptive features of the meters; (2) a holiday dataset summarizing the public holidays; and (3) corresponding hourly weather data. The "values" in the dataset present meter readings used to measure aggregate power consumption for the facility. Table II presents the software and hardware used to develop the system.

TABLE II. HARDWARE AND SOFTWARE SPECIFICATIONS.

| Hardware | Software |
| --- | --- |
| Core (TM) i5-4570 CPU at 3.20 GHz, 16 GB of DDR4 SDRAM, Windows 11 64-bit, and Nvidia GeForce RTX 2070 | Python v3.10.12, Jupyter Notebook v5.3.0, Pandas v1.5.3, NumPy v1.23.5, Matplotlib v3.6.3, Google Cloud |

For power-related data, the dataset is joined on the meter ID; for weather and holiday data, the dataset is joined on the date stamp. This work focuses on power usage for the meter located at Site 38. The missing data were handled using multiple imputation by chained equation (MICE) imputation. MICE imputes the missing data through an iterative series of predictive models. In each iteration, each variable is imputed using the other variables in the dataset, and the iterations continue until convergence has been met.

The dataset generated has 49,000 rows. To address irregular time gaps, the differences in timestamps were calculated, which is the duration between consecutive measurements. Moreover, the difference in meter readings between consecutive timestamps provided the consumption data. Fig. 1 displays the consumption plot based on average hourly consumption, aggregated average weekly consumption, and average monthly consumption.

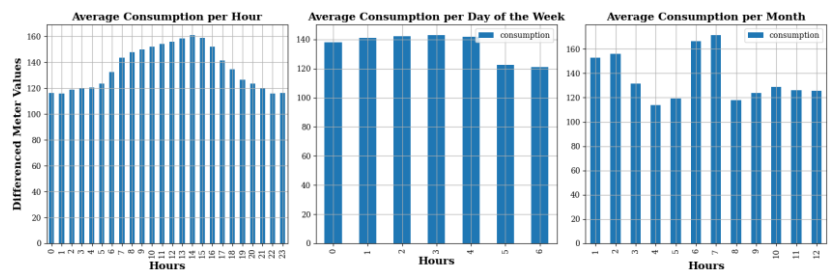

Fig. 1. Power consumption based on differences in meter readings.

Consumption can be seen at its peak during the middle to the afternoon of the day; consumption decreases during weekends, which indicates commercial buildings. Consumption is also high during the beginning months of the year (January and February) and at its peak in June and July. These are normal patterns considering the winter season at the beginning of the year and the summertime during June and July. Moreover, the difference not only makes the series stationary, but the presence of anomalies can also be observed through spikes in the stationary series (Fig. 2).

---

[4] https://shorturl.at/bqrM9

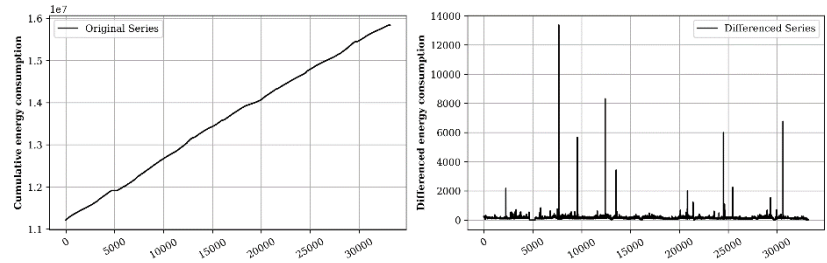

Fig. 2. Cumulative and differenced values of meter readings.

Environmental factors such as climate (temperature) affect power consumption by some equipment, such as heating, ventilation, and air conditioning (HVAC) systems. Fig. 3 displays the consumption pattern versus the temperature.

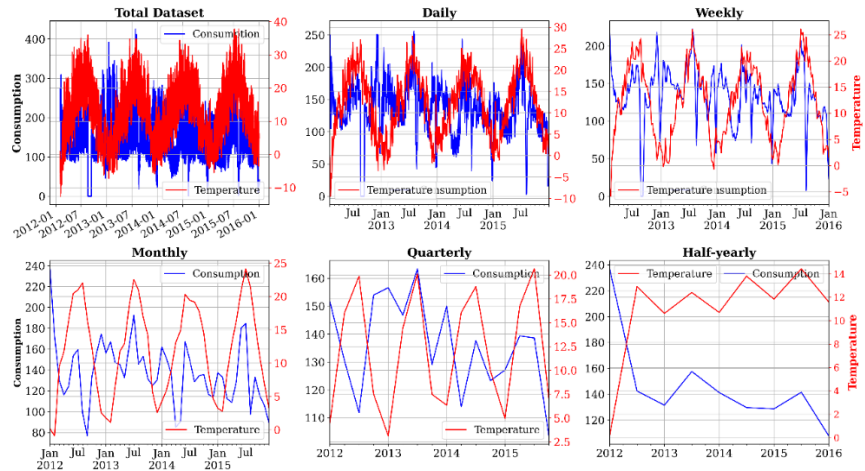

Fig. 3. Power consumption versus temperature over time.

Fig. 4 provides the visibility of consumption on aggregated data (daily, weekly, monthly, quarterly, and half yearly). The consumption is low during the early hours of the day, at its peak during the middle of the day, and slows down late in the evening. Also, a higher consumption pattern can be seen during weekdays and June and July. These are regular consumption behaviors, in line with our understanding.

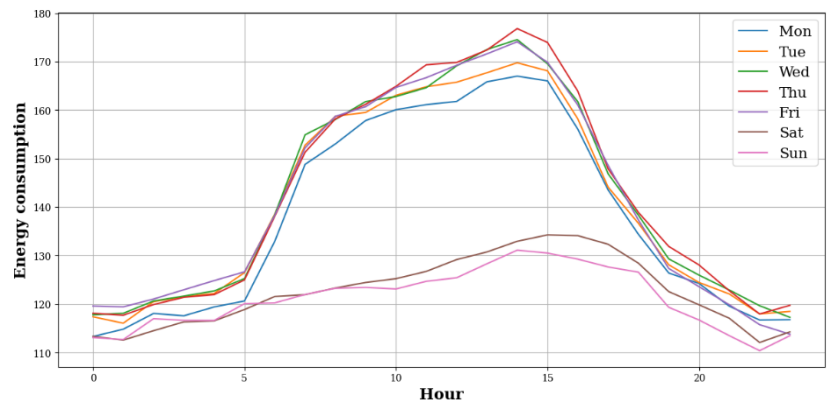

Fig 4. Weekly power consumption pattern.

The box plots in Fig. 5 display the distribution of consumption over different time intervals (days of the week, months of the year, hours of the day, and years). This helps us understand the temporal dynamics of the data. The anomalies beyond the whiskers are based on the interquartile range (IQR), which is the natural variability within the data. Although they are unusual data points, they may still be valid observations and are not necessarily indicative of anomalies. Power consumption patterns often vary over time, exhibiting daily, weekly, seasonal, and even yearly trends. Here, the data are multivariate, meaning that anomalies may arise from complex interactions and relationships between multiple factors, which may not be fully captured by univariate outlier detection methods. Therefore, we decided to use multivariate model training.

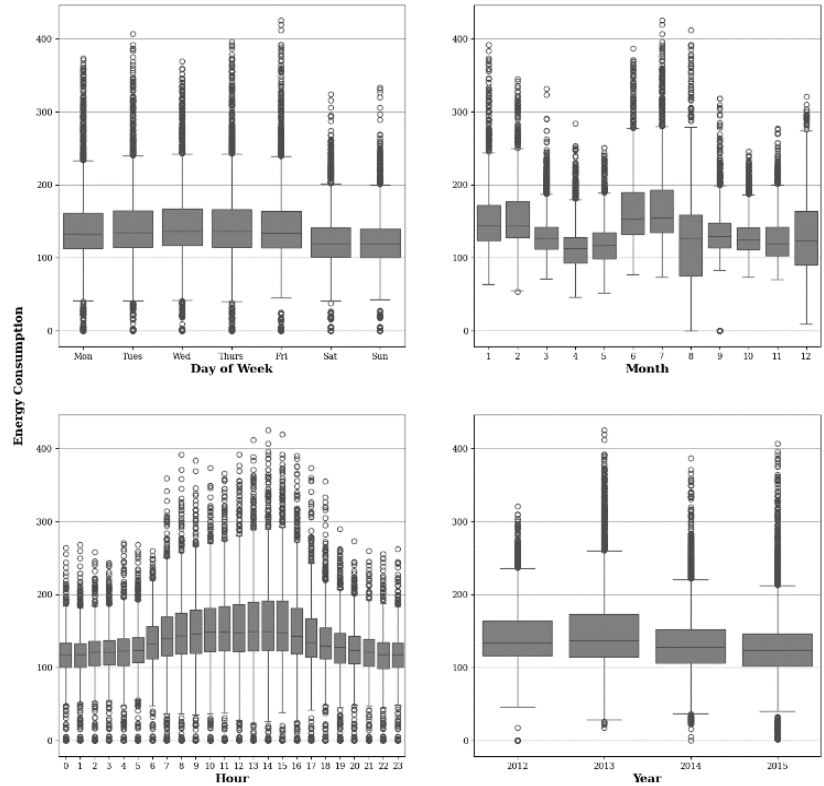

Fig 5.  Consumption distribution across time intervals.

The feature engineering was performed on the explanatory variables, which included the first- and second-lag values of consumption, along with a daily and monthly lag that captures the multiple levels of seasonality. It was found that fewer than 500 observations were not recorded every hour, which is an error in data entry. However, these values can still be a useful approximation of the month/weekday shift.

Fig. 5 presents the Pearson correlation matrix. We can see that dayshift is significantly correlated with consumption, which makes logical sense.

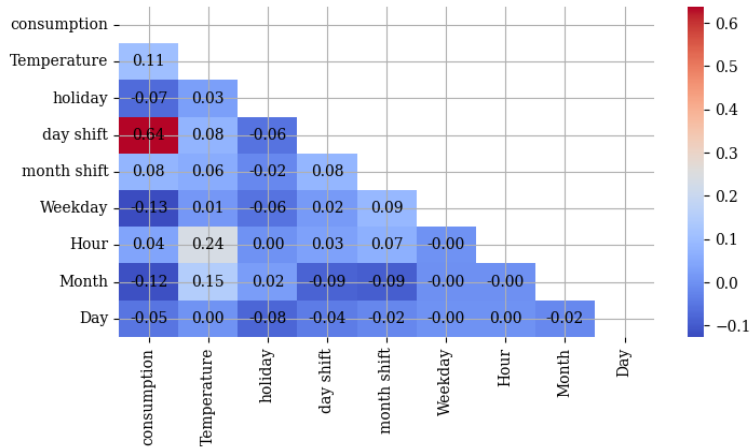

Fig. 6. Correlation plot.

Table III presents the statistical summary of the final list of columns, including the consumption (target column) and the other explanatory variables. The mean consumption is approximately 135.68 units, with a standard deviation of 50.08 units indicating variability in consumption levels. This is approximately 36.85% of the mean consumption value (50.08 / 135.68) , indicating moderate variability. The maximum consumption value is 425 units. This looks natural and can arise due to fluctuations in demand, changes in weather patterns, or periodic events.

TABLE III. STATISTICAL SUMMARY.

| index | consumption | temperature | holiday | Month shift | week day | hour | month | day |
|---|---|---|---|---|---|---|---|---|
| count | Number of data points: 30,000 | | | | | | | |
| $\mu$ | 137.98 | 12.52 | 0.02 | 139.60 | 3.00 | 11.49 | 6.20 | 15.57 |
| $\sigma$ | 51.02 | 8.44 | 0.17 | 65.01 | 2.00 | 6.92 | 3.27 | 8.75 |
| min | 0.00 | -12.60 | 0.00 | 0.00 | 0.00 | 0.00 | 1.00 | 1.00 |
| max | 425.00 | 37.70 | 1.00 | 425.0 | 6.00 | 23.0 | 12.0 | 31.0 |
| 25% | 110.00 | 5.90 | 0.00 | 113.0 | 1.00 | 5.00 | 3.00 | 8.00 |
| 75% | 161.00 | 18.60 | 0.00 | 163.0 | 5.00 | 17.0 | 9.0 | 23.0 |
| median | 130.00 | 12.40 | 0.00 | 132.0 | 3.00 | 11.0 | 7.00 | 16.0 |

Sudden changes in 1-lag, 2-lag consumption, day shifts, and month shifts (e.g., large σ, extreme minimum or maximum values) indicate important shifts or anomalies in the data. Table IV displays the statistical results. After removing the extreme values, the statistical distribution of the data became more normal. Fixing the data error led to minor adjustments in the mean while not significantly affecting other statistical measures.

TABLE IV. STATISTICAL SUMMARY.

| index | 1-lag | 2-lags | day shift | month shift |
|---|---|---|---|---|
| count | Number of datapoints: 33167 | | | |
| $\mu$ | 135.53 | 135.69 | 135.81 | 137.54 |
| $\sigma$ | 50.06 | 50.06 | 50.04 | 49.67 |
| min | 0.00 | 0.00 | 0.00 | 0.00 |
| max | 425.00 | 425.00 | 425.00 | 425.00 |
| 25% | 110.00 | 110.00 | 110.00 | 111.00 |
| 75% | 158.00 | 158.00 | 158.00 | 160.00 |
| median | 130.00 | 130.00 | 130.00 | 132.00 |

We performed the nonparametric Kolmogorov–Smirnov goodness-of-fit test and the Anderson–Darling test. Table V displays the statistics showing that the data are not normally distributed. The Kolmogorov–Smirnov test is more sensitive to the distribution center, while the Anderson–Darling test is far more sensitive to the distribution's tails. The analysis provides a detailed understanding of consumption patterns, identifies potential anomalies, and lays the groundwork for further anomaly detection and modeling efforts.

TABLE V. NORMALITY TEST.

| Kolmogorov–Smirnov Test | | |
|---|---|---|
| Statistic: 0.980 | p value: 0.0 | The data do not appear to be normally distributed. |
| **Anderson–Darling test** | | |
| Statistic: 543.647 | Critical values: [0.576 0.656 0.787 0.918 1.092] Significance levels: [15. 10. 5. 2.5. 1] | The data do not appear to be normally distributed at the 5% significance level. |

Fig. 5 displays the skewness and kurtosis of the consumption data. The positive skewness of the distribution is skewed to the right, meaning that the tail on the right side of the distribution is longer or fatter than that on the left side. This suggests that there may be some periods of relatively higher consumption compared to the overall average, leading to a right-leaning distribution. A kurtosis of 2.78 suggests that the distribution is slightly leptokurtic, meaning that it has slightly heavier tails and is more peaked compared to a normal distribution.

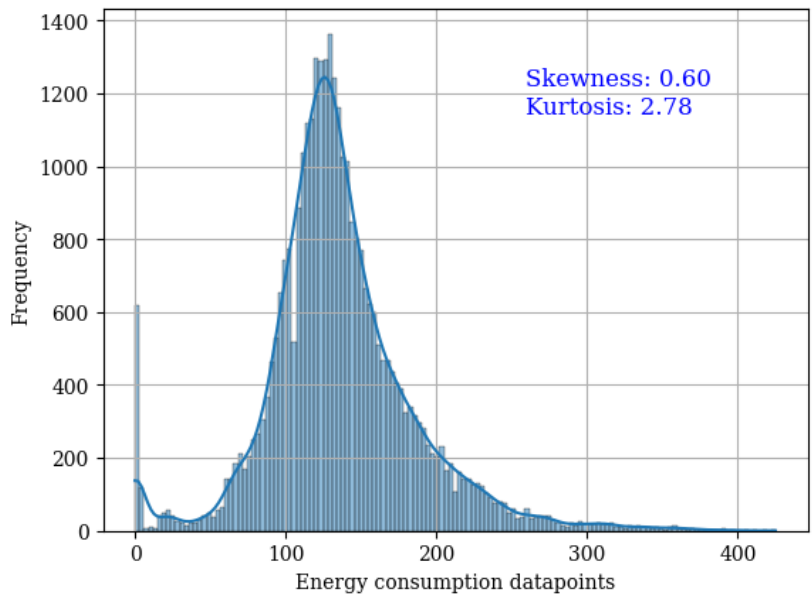

Fig. 7.  Skewness and kurtosis analysis of consumption data.

The final dataset has 48,995 rows after necessary cleaning and 10 columns, including the power consumption (difference in meter reading). It is further split into three different sets: the training set [:30000], the test set [30000:40000], and the validation set [40000:].

### III. METHODOLOGY

This study investigates both supervised and unsupervised learning processes. Omar et al. presented an overview of the research directions for supervised and unsupervised methods for anomaly detection (Omar et al., 2013). Although their research has brought to light various benefits and drawbacks, technology has advanced significantly since then, and the fields of artificial intelligence and machine learning are constantly changing.

#### A. Supervised learning

Theoretically, supervised anomaly detection is superior in its overall accuracy due to the clear understanding of normality and abnormality (Shaukat et al., 2021). We first experimented with a supervised Light Gradient Boosting Machine (GBM). It uses a histogram-based algorithm, i.e., it buckets continuous feature values into discrete bins, which accelerates the training procedure and results in lower memory usage. Moreover, since it is a tree structure, the scale of the features does not matter here. This is because at each tree level, the score of a split will be equal whether the respective feature has been scaled or not.

Park et al. (2021) introduced a sliding window-based LightGBM for anomaly detection. Although sliding window techniques have advantages in capturing temporal dependencies and adapting to changes in data distribution over time, they may not be optimal for real-time anomaly detection in rapidly changing environments. Instead, we used a rolling window, which is beneficial for near-real-time anomaly detection. By continuously updating and analyzing data within a fixed window size (24 hours) as new data points become available, rolling window methods enable the detection of anomalies with minimal delay.

Using threshold-based monitoring is an easy way to detect anomalies. With this method, an alert is set off when the energy consumption surpasses a predetermined threshold. However, this method struggles to keep up with the changing energy use patterns of buildings (Rashid & Singh, 2018). According to Balakrishnan et al., this strategy produced more than 10,000 alarms every day on the University of California San Diego campus (Narayanaswamy et al., 2014). The goal of finding anomalies was defeated when the building authorities disregarded the large number of alarms. Therefore, it is critical to create a dynamic method that can adjust to changing energy demand and has a low false positive alert rate. Current techniques consider daily data prior to identifying an anomaly on a certain day. However, in doing so, these approaches treat every day the same and disregard patterns of spending.

We used the Mahalanobis distance for each data point, considering the covariance structure among variables. It detects outliers based on the pattern of data points by utilizing the inverse of a covariance matrix of variables in conjunction with the center of the data. This is a multivariate outlier detection strategy since it can locate distances in multidimensional spaces. It also captures complex patterns and relationships that may not be apparent in univariate analyses.

The Mahalanobis distance ($D_M$) for each predicted value from the supervised model is calculated using Eq. (1):

$$D_M(x, \mu, \Sigma) = \sqrt{(X - \mu)^T \Sigma^{-1} (x - \mu)} \quad (1)$$

where $x$ = the predicted value, $\mu$ = the mean vector of the features in the validation data, and $\Sigma$ = the covariance matrix of the features in the validation data.

The moving averages of the Mahalanobis distances are formulated using the convolution operation with a window size $w$:

$$(MAV)_i = \frac{1}{w} \sum_{j=0}^{w-1} D_M (i + j) \quad (2)$$

Dynamic thresholds for anomalies are determined based on the 95[th] percentile of $D_M$ and moving averages $MAV_i$. Anomalies are classified using Eqs. (3), (4) and (5):

Anomalies based on the Mahalanobis distance:

$$\begin{cases} 1, & if\ D_M[i] > MahalanobisThreshold \\ 0, & otherwise \end{cases} \quad (3)$$

Anomalies based on the moving average:

$$\begin{cases} 1, & if\ MAV[i] > MoveAverageThreshold \\ 0, & otherwise \end{cases} \quad (4)$$

Anomalies based on the combined score:

$$\begin{cases} 1, & if\ MAV[i] > MoveAverageThreshold \\ 0, & otherwise \end{cases} \quad (5)$$

While the combined approach provides a unified metric, it misses some anomalies compared to using only the Mahalanobis distance, as displayed. Fig. 8 displays the anomalous points based on the combined score. This means that if either the Mahalanobis distance or the moving average is below the threshold, the combined approach will not classify it as an anomaly.

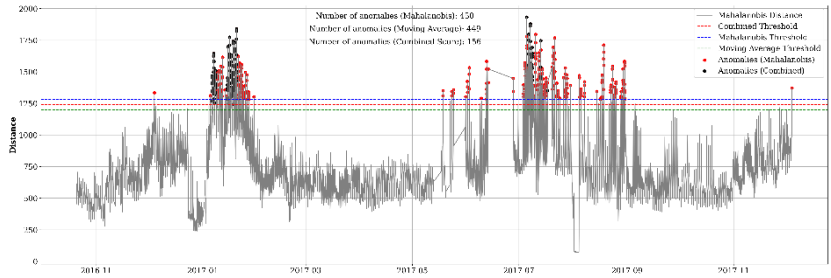

Fig. 8. Anomalies in the validation set where the dynamic threshold was based on the Mohalanomis distance, simple moving averages, and combined score.

Fig. 9 displays the anomaly detection using the Mahalanobis distance with the threshold lines and detected anomalies. We used 5-fold cross-validation to evaluate the performance of each fold. In this way, it detects 500 anomalies on the test set.

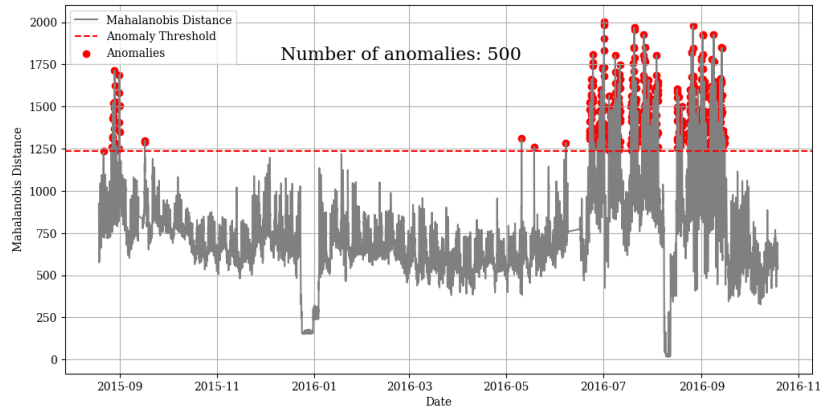

Fig. 9. Anomalies on the test set based on the Mohalanomis distance.

Fig. 10 displays the decision tree plots, showing how the decisions are made by the model and how the features interact with each other. $prev\ value$ is identified as the root node. The anomalous decision is made when $prev\_value > 154.33$. The subsequent splits, such as $prev\_value \leq 211.50$ and $prev\_value \leq 274.5$, further refine this partitioning process. The splits enable the tree to capture complex relationships and patterns in the data. Here, the tree mostly leverages the $prev\ value$ feature to make predictions about consumption, and subsequently, the threshold determines the anomalous points.

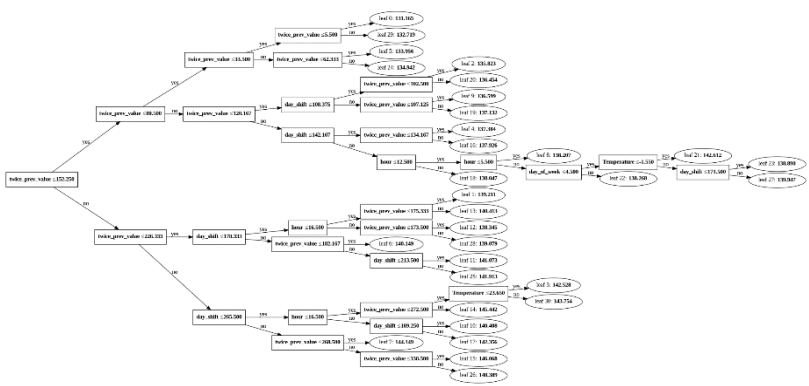

Fig. 10. Decision tree plot.

Fig. 11 displays the "*feature importance*" plot for the test dataset. Despite having multiple variables, the model heavily relies on previous values and two previous values, suggesting that temporal patterns and trends are crucial in predicting the target variable. $day\ shift$ It also predicts the target variable, albeit less than the previous two features. $hour$ contributes slightly to the model's predictions. The $Temperature$ variable has relatively lower importance but still provides some predictive value. The $day\ of\ week$ variable has minor importance, indicating that certain days may have a slight impact on consumption. Likewise, the $month$ feature contributes minimally, and $month\ shift$ has the lowest importance. The $holiday$ variable has negligible importance, indicating that holidays are not significantly influential in this context.

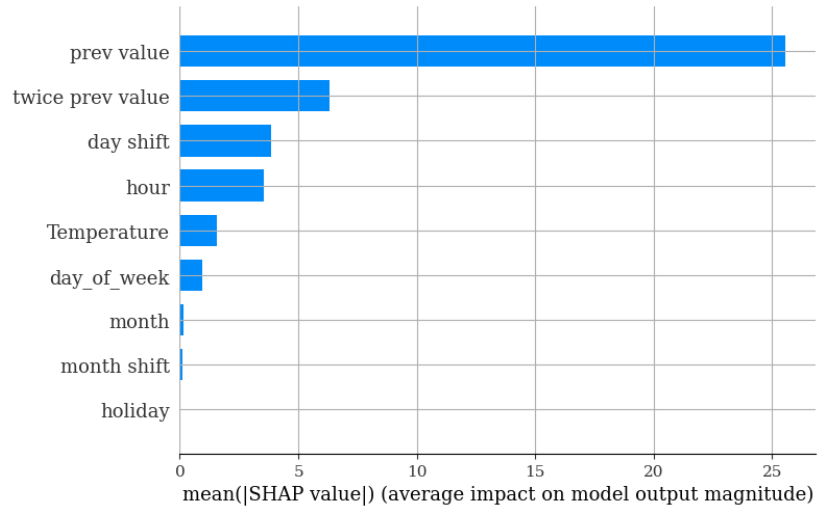

Fig. 11. Important features for decision making.

We further employed hierarchical clustering to enhance the interpretability of anomalies by organizing them into meaningful groups based on their characteristics. Fig. 12 displays the dendrogram from hierarchical clustering. It visualizes two key pieces of information: which points are grouped together and the similarity of members of the group. Groups are shown through horizontal lines; the Y position of the horizontal lines shows the distance between the group's members. This distance's value is a product of the distance function and the linkage (unweighted pair group method with arithmetic mean (UPGMA)). UPGMA calculates the distance between two groups as the average distance between each point of the first group and each point of the second group. The plot is shown with the X-axis as vertical with a timestamp and the Y-axis as horizontal (Euclidean distance).

The greater the distance of the vertical lines in the dendrogram is, the greater the distance between those clusters. Therefore, we set a threshold such that it cuts the tallest vertical line. The clusters represent groups of anomalies exhibiting similar behavior. Moreover, each cluster may correspond to a different type of anomaly; for example, one cluster might represent anomalies related to sudden spikes in power consumption, while another cluster might represent anomalies related to sustained deviations from expected power usage. Data points that do not fall within any of the identified clusters represent anomalies that do not conform to any specific pattern. These data need to be investigated further to understand the underlying causes. The average silhouette score is about 0.229, indicating the clusters are well separated. A higher silhouette score with values closer to 1 indicates dense, well-separated clusters. The average Davies–Bouldin index is approximately 1.273. This index measures the average "similarity" between each cluster and its most similar cluster, with lower values indicating better clustering. An index closer to 0 indicates better clustering. The silhouette score suggested moderate separation between clusters, while the Davies–Bouldin index indicated that the clusters were well defined.

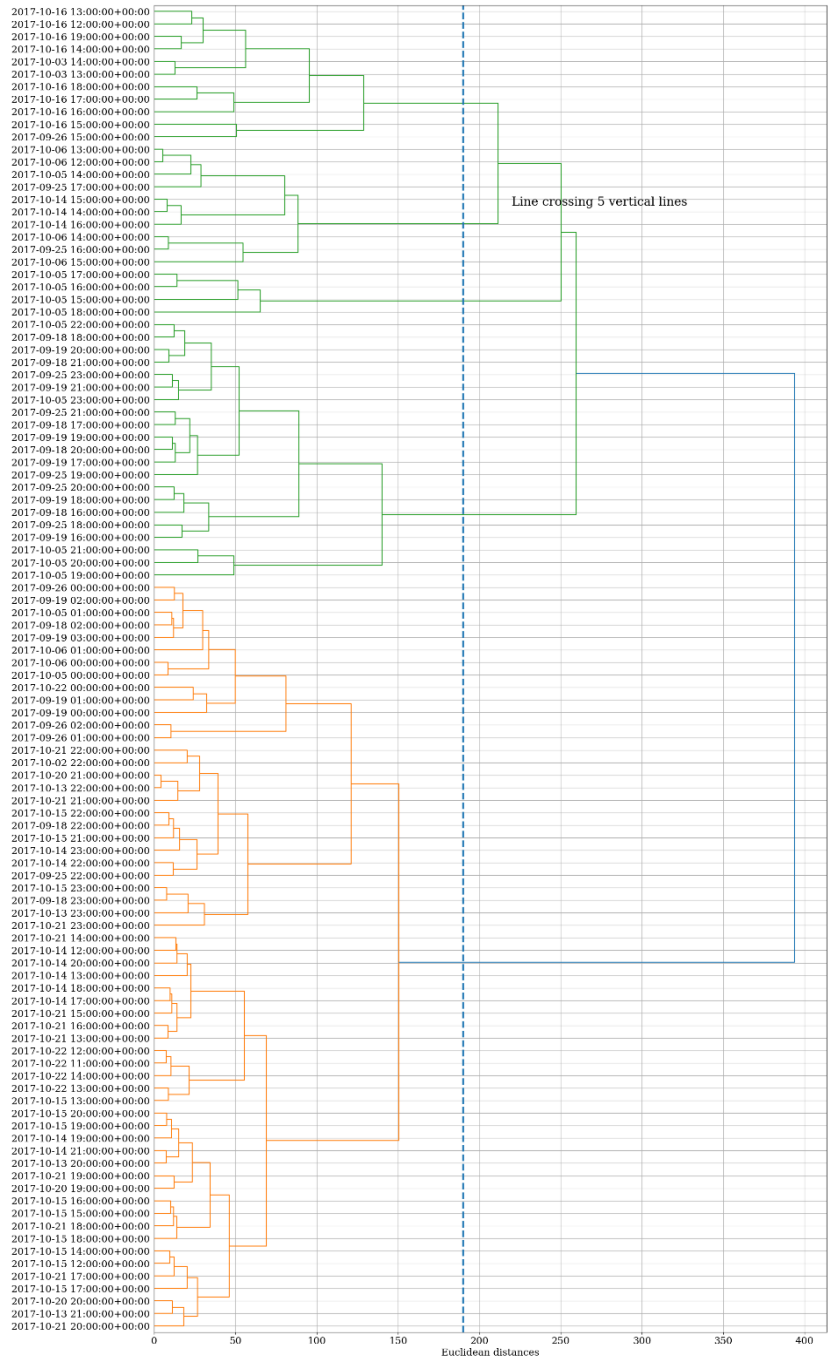

Fig. 12. Sliced dendrogram for hierarchical clustering.

ML models provide point estimates (predictions), but they often come with uncertainty due to sampling variability and model randomness. We used bootstrap resampling to estimate the uncertainty in the model's mean absolute error (MAE) by repeatedly resampling the dataset and calculating the metric for each sample. Bootstrap resampling does not assume any specific distribution for the data. It provides a nonparametric way to estimate uncertainty. The output provides a 95% confidence interval for the MAE of the model's predictions. This interval [10.27, 10.75] suggests that we can be 95% confident that the true MAE falls within this range. The interval is small, which indicates higher precision and confidence in the estimation of the MAE.

Studies suggest that supervised approaches are superior; however, a few real-world issues seriously impede their use (e.g., Himeur et al., 2021;

Shaukat et al., 2021). Despite these challenges, this work demonstrates the effectiveness of utilizing the Mahalanobis distance and rolling windows for near real-time anomaly detection. By leveraging dynamic thresholds and hierarchical clustering, this work provides insights into detecting and inferring anomalies, thereby facilitating proactive power management strategies.

*B. Unsupervised learning*

Recently, the scientific community has made significant strides, especially in the areas of machine learning (ML) and deep learning (DL). Both machine learning and deep learning models can be taught for different tasks and can learn exclusively from data. In certain application domains, accessible labeled data are restricted, and the amount of labeled data available is insufficient to train a model successfully. In anomaly detection, we must be prepared to address datasets containing minimally labeled data. A recent review and comparative evaluation of unsupervised anomaly detection methods (e.g., Liu et al., 2021; Nassif et al., 2021; Finke et al., 2021) focused on unsupervised learning.

The use of autoencoders in power consumption and management has recently attracted the attention of several researchers (see [19]; [20]; etc.). Autoencoders are effective at reconstructing data in general, especially in time series (see [21], [22]), and researchers have reported anomaly identification in power consumption series using autoencoders (see [23]; [24]; [25]). It is composed of two essential components: an encoder and a decoder. The input data are compressed by the encoder, a neural network, into a small number of latent space variables. Another network serves as the decoder, reconstructing the original data from the latent space variables. Both components are trained as a single network to reconstruct the input data as closely as possible by selecting an appropriate loss function.

*1) Model training steps*

Prior to modeling, we checked the data distribution of the original dataset to determine whether data normalization was necessary. The autoencoder uses the reconstruction error as the anomaly score. Points with a high score are anomalies. At inference, the autoencoder reconstructs the normal data. If the anomaly is too small and close to the normal data, the reconstruction error will be small, and thus, the anomaly will not be detected. To overcome this problem, the autoencoder is normalized to ensure that the input data contain no anomalies before performing a good reconstruction. Similar data splits (training, testing, and validation) are used here to ensure consistent training.

Figs. 14 and 15 display the data distribution before and after the data standardization. Standardization was performed using the MinMax scaler, which scales the features between 0 and 1 while preserving the shape of the original distribution. We observed that the data were slightly more uniform and proportionally distributed after normalization, and the ranges were also reduced to fit values between 0 and 1.

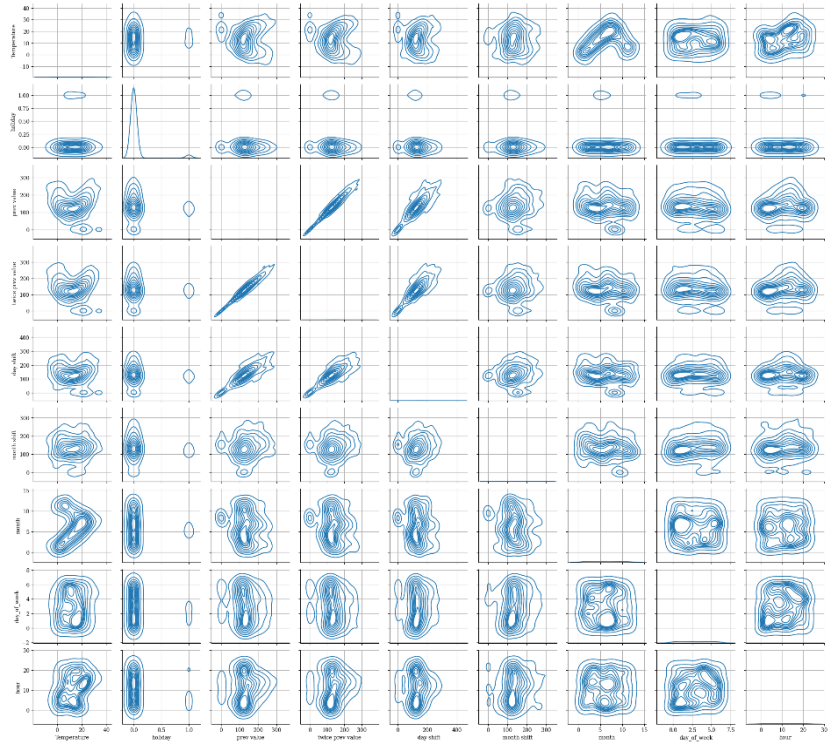

Fig. 13. Data distribution (original dataset)

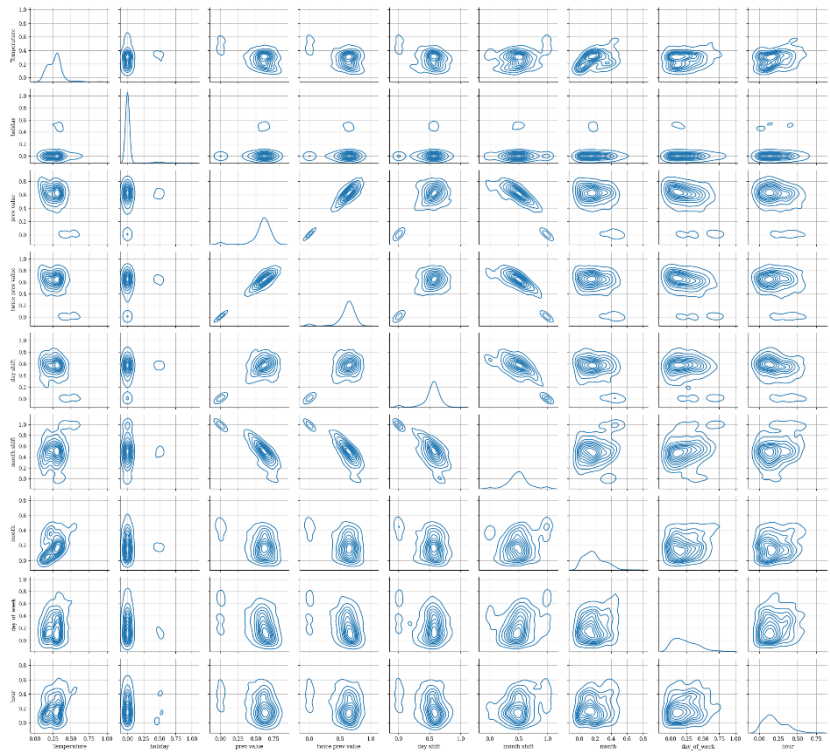

Fig. 14. Data distribution (standardized dataset)

We implemented a simple network architecture based on a fully connected layer using the exponential linear unit (ELM) activation function. The objective is to train a model to reconstruct normal consumption patterns and detect potential abnormalities in the data. Model training is performed with an automatic stopover function, where the training continues if the model is reducing the training loss. Finally, only the weights for the model with the lowest validation loss are saved. Table VI displays the model architecture with both the encoder and decoder.

TABLE VI. CNN-AUTOENCODER MODEL ARCHITECTURE

| Layer Type | Output Shape | Params | Activation Function |
|---|---|---|---|
| Deconstruct (encode) | | | |
| dense (Dense) | (None, 9) | 90 | elu |
| dense_1 (Dense) | (None, 16) | 160 | elu |
| dense_2 (Dense) | (None, 8) | 136 | elu |
| dense_3 (Dense) | (None, 4) | 36 | elu |
| dense_4 (Dense) | (None, 2) | 10 | elu |
| Reconstruct (decode) | | | |
| dense_5 (Dense) | (None, 4) | 12 | elu |
| dense_6 (Dense) | (None, 8) | 40 | elu |
| dense_7 (Dense) | (None, 16) | 144 | elu |
| dense_8 (Dense) | (None, 9) | 153 | elu |

The encoder compresses the input data into a lower-dimensional representation (latent space), and the decoder reconstructs the original input from this representation. The Adam optimizer is used for gradient descent. Fig. 16 displays the training and validation losses for different batch sizes (16, 32, 64, 128, and 256). Since the batch size did not make any difference in losses, we considered a batch size of 256 for the training process, and as the plot shows, the training is quite stable.

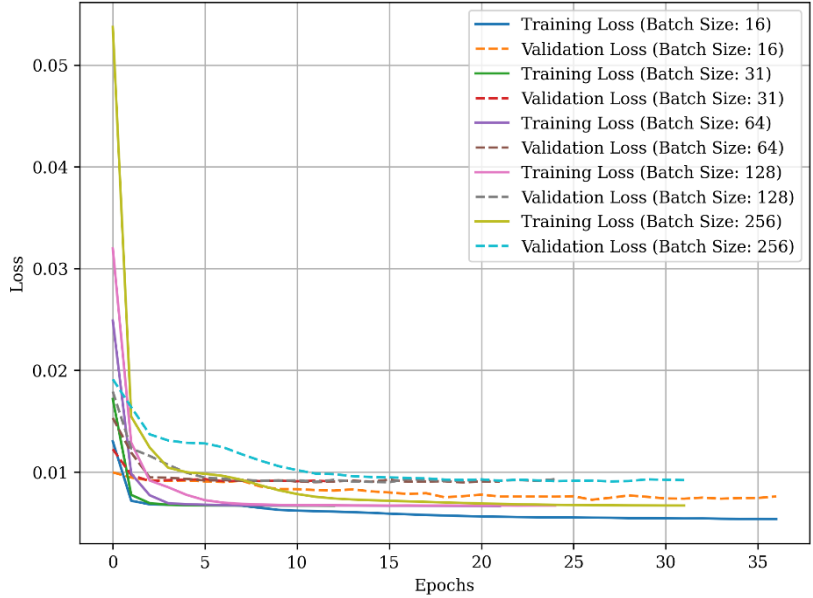

Fig. 15. Training and validation loss for different batch sizes

Fig. 17 displays the distribution of reconstruction errors, followed by Fig. 18, which displays the anomalous points. The distribution is concentrated toward 0 on the x-axis, which suggests that most of the data have minimal error. However, the data fall near 0.05 and beyond the x-axis, indicating significantly larger errors with potential anomalies. We set a threshold to limit the number of false positives to a manageable degree and capture the most anomalous data points. The dynamic threshold is applied based on the mean and standard deviation of the reconstruction errors displayed in Eq. 6.

$$DynamicThreshold = \mu(ReconstructionError) + k * \sigma(ReconstructationError) \quad (6)$$

where $\mu$ = average, $k$ = multiplier and $\sigma$ = standard deviation. $k$ is a user-defined number that adjusts the sensitivity of the threshold, as displayed in

Fig. 17. Eqn. 7 and 8 formulate the reconstruction error and subsequent anomaly detection, respectively.

$$ReconstructionError = \frac{1}{n}\sum_{i=1}^{N}(y_i - y'_i)^2 \quad (7)$$

$$Anomaly = \begin{cases} 1, & if ReconstructionError > DynamicThreshold \\ 0, & otherwise \end{cases} \quad (8)$$

where $N$ is the number of data points, $y_i$ and $y'_i$ are the $i^{th}$ data points in the original and reconstructed sequences, respectively, $x$ is a data point in the dataset, and the multiplier $k$ is used to adjust the sensitivity of the threshold.

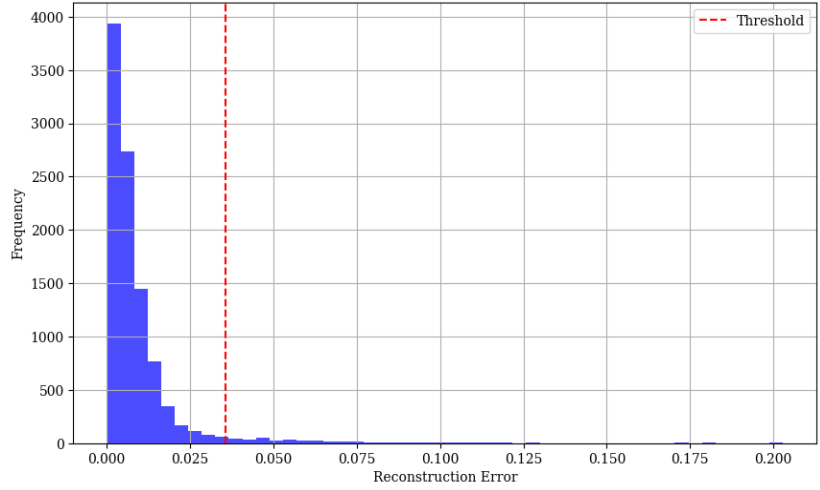

Fig. 16    Distribution of reconstruction errors.

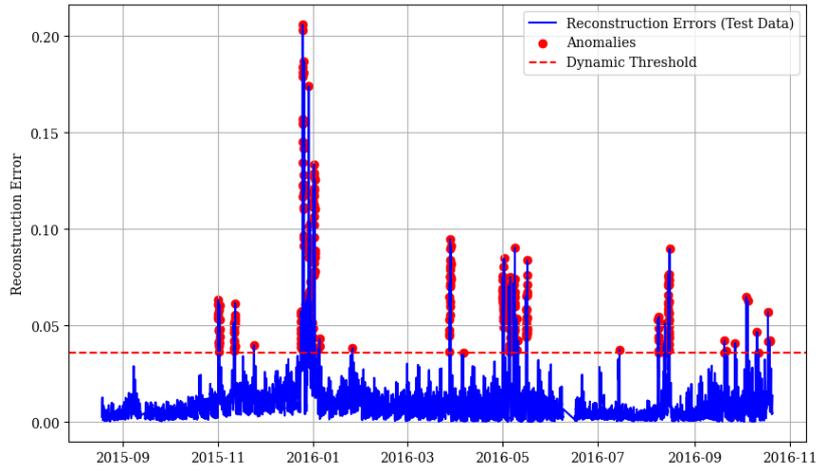

Fig. 17    Anomalies on the test dataset with a dynamic threshold.

The model detected 360 anomalies from the test dataset (Fig. 18). Finally, the model is applied to a hold-out dataset, where 385 anomalies are detected (Fig. 19). The autoencoder detected fewer anomalies than the supervised LightGBM model in both the test and validation split (Table VII). However, we can observe some interesting spread-out anomalies with the autoencoder during 2016–11 and 2017–05 compared to LightGBM, which is more concentrated in a certain specific time range.

Table VII. CNN-Autoencoder Model Architecture

| Approach | Test | Validation |
|---|---|---|
| Supervised (LightGBM) | 500 | 450 |
| Unsupervised (Autoencoder) | 345 | 385 |

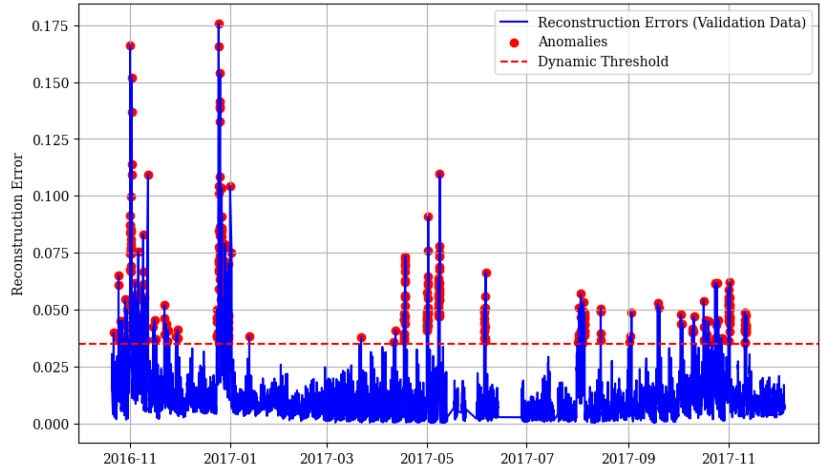

Fig. 18.  Anomalies on the hold-out dataset.

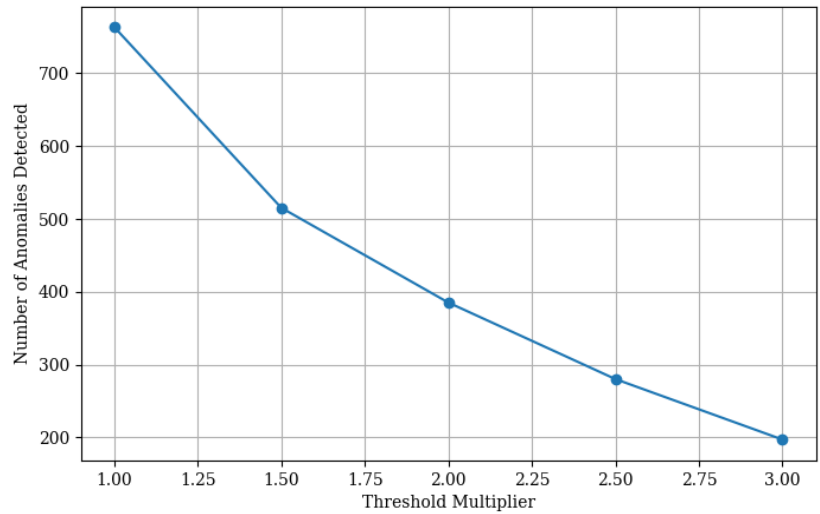

Fig. 19.  Number of anomalies detected vs. threshold multiplier.

The proposed approach is close to real-life deployment, which makes it well suited for deployment. It is interpretable and displays a timestamp for immediate human supervision. Alert and visualization are effective methods for immediately discovering probable anomalies in the data.

To summarize, this section demonstrates the effectiveness of unsupervised anomaly detection using an autoencoder for power consumption data. By leveraging reconstruction errors and dynamic thresholding, the proposed approach offers a data-driven solution for detecting anomalies without the need for labeled data. While fewer anomalies may be detected compared to supervised approaches, the observed spread-out anomaly patterns contribute to a comprehensive understanding of power consumption dynamics.

## IV.  LIMITATIONS AND FUTURE DIRECTIONS

The basis of anomaly identification requires stable and accurate consumption data. The data inconsistencies, missing values, or noise affect the accuracy of a model. Therefore, data mining and ETL require specialized skills prior to modeling. Moreover, the proposed system combines statistical methods with both supervised and unsupervised approaches. While supervised tree-based models may be easy to explain, unsupervised approaches require in-depth technical skill. Understanding the root cause of anomalies and taking necessary actions based on them may require further investigation and domain expertise. The dynamic thresholds based on moving averages and MD may require fine-tuning and adjustment for different datasets. Selecting an appropriate threshold that balances false positives and false negatives can be challenging. These limitations would require further research and refinement of the proposed methodology, potentially exploring hybrid approaches combining different ML techniques or leveraging additional domain knowledge. Moreover, aggregated-level consumption data may not be the best way to detect anomalies in power consumption (see Rashid et al., 2019; Himeur et al., 2021) because they cannot provide accurate information on the causes. Therefore, the proposed model needs to be tested at the appliance level to measure its efficacy and make the system more attractive for real-life usage.

It is advisable to use locally weighted scatterplot smoothing (LOWESS) instead of moving averages. LOWESS provides a smoothed curve by fitting localized regression models to subsets of the data, which can result in a smoother representation of the underlying trend compared to simple moving averages. Furthermore, most power consumption anomaly detection solutions do not suffer from security attack resistance resulting from inadvertently incorrect system development. Moreover, future research may explore hybrid approaches that combine supervised and unsupervised methods to enhance anomaly detection accuracy in dynamic environments.

## V. Conclusion

This study investigated anomaly detection, which has yielded valuable insights and advancements. The proposed approach provides an integrated framework for anomaly detection and promotes data-driven decision-making by combining visual analytics with decision science principles. Leveraging unsupervised machine learning techniques, this work demonstrates a robust methodology for detecting anomalies in power consumption patterns. Through the reconstruction of normal consumption patterns, anomalies are effectively flagged, allowing timely intervention and power management strategies. Moreover, the implementation of a dynamic threshold adds a layer of flexibility and adaptability to the detection process. By adjusting the sensitivity of the threshold based on the mean and standard deviation of reconstruction errors, this approach strikes a balance between sensitivity and specificity, minimizing false positives while capturing significant anomalies. Comparison with supervised machine learning using Light Gradient Boosting Machines, has provided valuable insights into the trade-offs inherent in different anomaly detection methodologies. While supervised methods may detect more anomalies and often provide better accuracy, unsupervised approaches offer advantages in scenarios where labeled data are not available. Moreover, the interpretability and immediate applicability of the real-life deployment of the proposed approach are emphasized. Since power management is becoming increasingly critical in the face of environmental and economic challenges, this work contributes significantly to optimizing power efficiency and sustainability. Future research in this direction could further explore hybrid approaches combining supervised and unsupervised methods to enhance anomaly detection accuracy and applicability across diverse power consumption scenarios.